\documentclass[runningheads]{llncs}

\usepackage[mobile]{eccv}

\usepackage{eccvabbrv}

\usepackage{graphicx}
\usepackage{booktabs}
\usepackage{multirow}

\usepackage[accsupp]{axessibility}  

\usepackage[pagebackref,breaklinks,colorlinks,citecolor=eccvblue]{hyperref}

\usepackage{orcidlink}
\usepackage{colortbl}

\begin{document}

\title{ECHO: Towards Emotionally Appropriate and Contextually Aware Interactive Head Generation}

\titlerunning{ECHO}

\author{Xiangyu Kong\inst{1,2} \and
Xiaoyu Jin\inst{2} \and
Yihan Pan\inst{2} \and
Haoqin Sun\inst{3} \and
Hengde Zhu\inst{4} \and
Xiaoming Xu\inst{2} \and
Xiaoming Wei\inst{2} \and
Lu Liu\inst{1} \and
Siyang Song\inst{1}
}

\authorrunning{Xiangyu Kong et al.}

\institute{University of Exeter, Exeter, UK \and
Meituan, China \and
College of Computer Science, Nankai University, China \and
School of Computer Science and Technology, Tongji University, China\\
\email{\{xk219,S.Song,L.Liu3\}@exeter.ac.uk}
}
\maketitle

\begin{abstract}
In natural face-to-face interaction, participants seamlessly alternate between speaking and listening, producing facial behaviors (FBs) that are finely informed by long-range context and naturally exhibit contextual appropriateness and emotional rationality. Interactive Head Generation (IHG) aims to synthesize lifelike avatar head video emulating such capabilities. Existing IHG methods typically condition on dual-track signals (i.e., human user’s behaviors and pre-defined audio for avatar) within a \textit{short temporal window}, jointly driving generation of avatar's audio-aligned lip articulation and non-verbal FBs. However, two main challenges persist in these methods: (i) the reliance on short-clip behavioral cues without long-range contextual modeling leads them to produce facial behaviors lacking contextual appropriateness; and (ii) the entangled, role-agnostic fusion of dual-track signals empirically introduces cross-signal interference, potentially compromising lip-region synchronization during speaking. To this end, we propose \textbf{ECHO}, a novel IHG framework comprising two key components: a \textit{Long-range Contextual Understanding} (LCU) component that facilitates contextual understanding of both behavior-grounded dynamics and linguistic-driven affective semantics to promote contextual appropriateness and emotional rationality of synthesized avatar FBs; and a block-wise \textit{Spatial-aware Decoupled Cross-attention Modulation} (SDCM) module, that preserves self-audio-driven lip articulation while adaptively integrating user contextual behavioral cues for non-lip facial regions, complemented by our designed two-stage training paradigm, to jointly enhance lip synchronization and visual fidelity. Extensive experiments demonstrate the effectiveness of proposed components and ECHO’s superior IHG performance.

\keywords{Interactive Head Generation \and Emotionally Appropriate Facial Behaviors \and Long-range Contextual Understanding}
\end{abstract}

\section{Introduction}
\label{sec:intro}

Dyadic human-avatar interaction forms a foundation for human-like virtual agents and digital humans \cite{zhou2025interactive,zhou2022responsive}, aiming to emulate fluid turn-taking and mutual responsiveness of real face-to-face human conversations. In natural dyadic conversations, each participant continuously alternates between speaking and listening, producing facial behaviors (FBs) including speech-aligned lip motions \cite{zhu2025infp}, facial expressions and head movements \cite{song2025react,peng2025dualtalk,liu2024customlistener}, actively influencing the flow of conversation. Motivated by this need, IHG approaches have emerged, aiming to synthesize realistic avatar head videos for human-avatar interaction, enabling seamless switching between speaking and listening roles.

\begin{figure}[t]
  \centering
  \includegraphics[width=\linewidth]{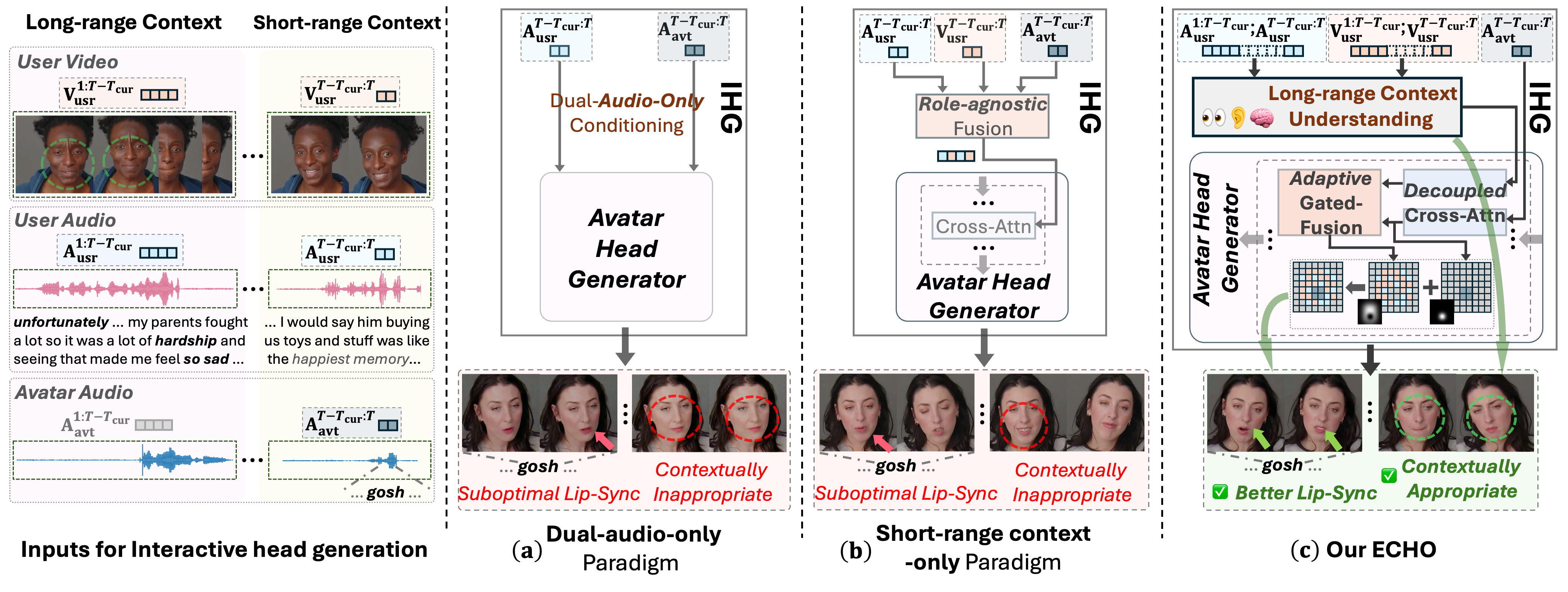}
  \caption{Overview of ECHO and pipeline comparison with existing IHG approaches. (a) Approaches \cite{zhu2025infp,cai2025towards,sun2025streamavatar} that rely exclusively on dual-track audio signals for avatar FBs generation; (b) Approaches \cite{peng2025dualtalk,ki2026avatar} that condition avatar FBs generation on short-range-only context modeling via role-agnostic fusion; (c) Our ECHO, which performs long-range contextual understanding and region-wise decoupled cross-attention mechanism, achieving contextually appropriate and precisely lip-synchronized avatar FBs.}
  \label{fig:overview}
  \vspace{-0.2cm}
\end{figure}

Despite growing progress, generating contextually appropriate and emotionally expressive avatar FBs still remains a central challenge in IHG. A large part of existing approaches \cite{zhu2025infp,cai2025towards,sun2025streamavatar} relies exclusively on \textit{dual-track audio behaviors} (i.e., expressed by human user and pre-defined for avatar) as driving signals, while ignoring human user's visual behaviors, to generate avatar's interactive FBs. However, this audio-only paradigm (Fig. \ref{fig:overview}-a) contradicts established findings \cite{boker2009effects,zhu2024perfrdiff,ng2023can}, suggesting that individuals' facial expressions and head movements also provide crucial non-verbal social-affective cues in eliciting their conversation partners' facial behaviors through mechanisms including emotional contagion and facial mimicry \cite{hatfield1993emotional,chartrand1999chameleon}. Consequently, this paradigm would theoretically degrade the diversity and contextual appropriateness of synthesized avatar facial behaviors. 
To bridge this gap, several recent studies \cite{peng2025dualtalk,ki2026avatar} incorporate speaker visual behaviors as complementary conditioning signals to guide the generation of avatar's facial behaviors. Nevertheless, conditioning signals in these methods typically scan only the current short temporal window, ignoring long-range context (Fig. \ref{fig:overview}-b) that is critical for semantic understanding of conversational dynamics and appropriate non-verbal response modeling \cite{zhu2024perfrdiff,ng2023can,luo2024reactface}. Consequently, the absence of such contextual understanding causes models to produce facial behaviors with context-inconsistency (e.g., responding with a smile in a situation characterized by sadness).

Beyond generating contextually appropriate and emotionally expressive facial behaviors, producing precise lip synchronization during speech responses is another fundamental requirement for IHG for dyadic interaction. To achieve this, existing IHG methods \cite{ki2026avatar,sun2025streamavatar,peng2025dualtalk} frequently inject dual-track input signals via entangled, role-agnostic fusion, i.e., they first fuse (e.g., via additive or attention-based fusion) the user's audio-visual behaviors and the pre-defined \textit{avatar self-audio} into a unified conditioning representation, which is then injected into the avatar head network (e.g., through cross-attention layer). However, such strategies may compromise lip synchronization quality (e.g., as suggested by Sync-scores \cite{chung2016out} metrics and qualitative visualizations), empirically indicating that the entangled dual-track fusion introduces cross-signal interference, particularly during speaking turns in the lip region, where articulation is expected to be predominantly governed by the avatar's own audio to ensure accurate audio-visual synchronization \cite{ki2025float,xu2024hallo,cui2025hallo3}. 

In this work, we propose a novel IHG framework called \textbf{ECHO}, to address these challenges. First, we introduce a \textit{Long-range Contextual Understanding} (\textbf{LCU}) component, which endows model with long-range contextual understanding, encompassing both user behavior-grounded and linguistic-driven affective dimensions, to jointly promote contextually appropriate and emotionally rational avatar FBs. Specifically, given long-range user audio-visual behaviors, it captures their fine-grained spatio-temporal dynamics via low-level perception encoding, and incorporates high-level representations encapsulating behavioral context cues, distilled by a multimodal understanding model \cite{xu2025qwen2}, yielding a unified representation for guiding avatar FBs generation. Then, the dyadic linguistic dialogue is further leveraged for linguistic context understanding with subsequent affective reasoning (performed via the LLM \cite{guo2025deepseek}), resulting in a descriptive avatar emotional state for explicit affective guidance of FBs generation. Second, to improve lip synchronization under dual-track conditioning, we introduce a block-wise \textit{Spatial-aware Decoupled Cross-Attention Modulation} (\textbf{SDCM}) module in avatar generator. SDCM injects dual-track conditioning inputs via region-wise decoupled cross-attention mechanism: it drives lip-region dynamics predominantly with avatar self-audio to preserve faithful articulation, while for non-lip facial regions, applying two decoupled cross-attention streams (respectively attending to avatar audio and user behavioral cues) followed by adaptive gated-fusion to produce FBs appropriately responsive to user behaviors yet self-audio-coordinated. Complementarily, a two-stage training paradigm with our hierarchical feature alignment (HFA) strategy (Sec. \ref{subsec:training_strategy}) is proposed to further preserve talking fidelity while enhancing contextual listening generation.

The contributions of our proposed framework are summarized below:
\begin{itemize}
    \item In this paper, we propose \textbf{ECHO}, a novel method that achieves interactive avatar head video generation where listening segments exhibit emotionally appropriate facial behaviors and speaking segments maintain precise lip synchronization with high visual fidelity.

    \item ECHO comprises a LCU component for long-range contextual understanding grounded in user behavioral dynamics and dyadic linguistic cues to facilitate contextual appropriateness of generated FBs; a SDCM module for faithful self-audio-driven lip articulation with appropriate non-lip facial dynamics; and a two-stage training paradigm with hierarchical feature alignment (HFA) to preserve talking fidelity while enhancing interactive listening generation.

    \item Extensive experiments on both interactive and single-role (listening and talking) avatar head generation validate effectiveness of each proposed component and demonstrate ECHO's superiority over existing methods in emotional expressiveness, lip synchronization, and visual quality.
\end{itemize}

\section{Related Work}
\label{sec:related_work}

\paragraph{Talking Face Synthesis.}
Audio-driven talking face synthesis aims to generate realistic talking head videos from a single reference image and driving audio signals, providing a foundational single-role generation capability that IHG methods can further build upon for interactive scenarios. Early methods adopt a two-stage paradigm that first predicts intermediate parametric representations from audio and then renders the final video. MakeItTalk \cite{zhou2020makelttalk} disentangles audio content and speaker embedding to predict facial landmark displacements, followed by image-to-image translation for video synthesis. SadTalker \cite{zhang2023sadtalker} predicts 3DMM expression and head pose coefficients via separate audio-conditioned networks, then renders talking faces through a 3D-aware face renderer. While effective, such parametric representations may limit fine-grained visual expressiveness.
Recent approaches shift toward end-to-end diffusion-based generation. Hallo \cite{xu2024hallo} introduces a hierarchical audio-driven visual synthesis module within a UNet-based stable diffusion \cite{rombach2021highresolution} model, employing hierarchical cross-attention to separately control lip motion, facial expression, and head pose for portrait image animation. Hallo3 \cite{cui2025hallo3} extends this line by adopting the video diffusion transformer (DiT) \cite{peebles2023scalable} as backbone, incorporating an identity reference network with causal 3D VAE to maintain consistent facial identity of generated videos. \cite{jiang2025omnihuman,lin2025omnihuman} scales up DiT-based human animation through mixed-condition training with text, audio, and pose signals, and further introduces MLLM-guided cognitive simulation for semantically aware motion generation. These methods mainly focus on single-role talking face generation from a single audio track, without modeling the responsive facial behaviors that arise from dyadic interaction, which is the core objective of IHG.

\paragraph{Listening Head Generation in Dyadic Interaction.}
In dyadic conversations, the listener/agent avatar's facial behaviors (FBs) serve as continuous non-verbal feedback that actively shapes conversational dynamics \cite{song2025react,zhu2025perreactor,liu2024customlistener}, making FBs generation a critical capability for building believable interactive agents. L2L \cite{ng2022learning} employs a VQ-VAE with cross-modal transformer to autoregressively predict non-deterministic facial motion from user's audio and facial motions. RLHG \cite{zhou2022responsive} introduces the ViCo benchmark with a temporal decoder baseline that maps user-conditioned features to avatar's expression and head motion coefficients. DIM \cite{tran2024dim} leverages masked prediction over joint user-agent quantized codes for dyadic interaction modeling. CustomListener \cite{liu2024customlistener} introduces text-guided listener generation with a Static to Dynamic Portrait module that transforms user-provided text descriptions into speaker-coordinated dynamic portrait tokens for controllable listener motion generation. PerFRDiff \cite{zhu2024perfrdiff} applies a diffusion transformer with hypernetwork-generated personalized weight shifts for listener-specific facial behavior generation.

\paragraph{Interactive Head Generation.}
Recent interactive head generation models further require handling both speaking and listening roles within a unified framework. ARIG \cite{guo2025arig} formulates interactive generation as autoregressive diffusion with shared attention across dual-track dual-modal signals and conversation state prediction. DualTalk \cite{peng2025dualtalk} generates facial motions in blendshape coefficient space via a Dual-Speaker Interaction Module with multimodal cross-attention and Transformer encoder-decoder. However, these methods operate within parametric spaces (3DMM \cite{blanz2023morphable,danvevcek2022emoca} or motion coefficients \cite{guo2024liveportrait}) with subsequent rendering to video, potentially constraining the fine-grained visual expressiveness of generated videos. INFP \cite{zhu2025infp} learns a motion latent space with verbal and non-verbal memory banks, using dual-track audio conditioning (human user and agent avatar) for implicit role switching but without requiring user's visual cues. Avatar Forcing \cite{ki2026avatar} employs causal diffusion forcing in motion latent space with cascaded cross-attention fusion and DPO-based expressiveness enhancement. A common limitation across these methods is that they predominantly rely on short-clip behavioral cues without long-range contextual modeling, thus limiting the contextual appropriateness and emotional expressiveness of generated facial behaviors. Our proposed ECHO aims to address these limitations, generating lifelike, lip-synchronized, and emotionally appropriate interactive avatar videos, as detailed in Sec. \ref{sec:method}. 

\section{Methodology}
\label{sec:method}

\textbf{Overview.} 
Grounded in human-like face-to-face interaction which entails user's long-range audio-visual behaviors and user-avatar multi-turn linguistic dialogue, our novel IHG framework \textbf{ECHO} leverages this long-range context alongside avatar's self-audio to jointly synthesize a $T_{\text{cur}}$-frame avatar head video clip as:
\begin{equation}
    \mathbf{V}^{T-T_{\text{cur}}:T}_{\text{avt}}
    = \text{ECHO}\big(\mathbf{V}^{1:T}_{\text{usr}}, \mathbf{A}^{1:T}_{\text{usr}},
    \mathbf{A}^{T-T_{\text{cur}}:T}_{\text{avt}}, \mathcal{P}_{\text{hist}}, \mathbf{I}_{\text{ref}},
    \mathbf{V}^{T-T_{\text{cur}}-T_{\text{prev}}:T-T_{\text{cur}}}_{\text{avt}}\big),
\end{equation}
where $T$ is current temporal position, $\mathbf{A}^{1:T}_{\text{usr}}$ and $\mathbf{V}^{1:T}_{\text{usr}}$ denote user’s long-range audio-visual behaviors, $\mathbf{A}^{T-T_{\text{cur}}:T}_{\text{avt}}$ is the avatar’s self-audio, $\mathcal{P}_{\text{hist}}$ denotes multi-turn dyadic dialogue history encapsulating affective context for emotion guidance. The avatar reference image $\mathbf{I}_{\text{ref}}$ preserves identity, and $\mathbf{V}^{T-T_{\text{cur}}-T_{\text{prev}}:T-T_{\text{cur}}}_{\text{avt}}$ denotes preceding $T_{\text{prev}}$ generated avatar frames for temporal continuity.

\begin{figure}[t]
  \centering
  \includegraphics[width=\linewidth]{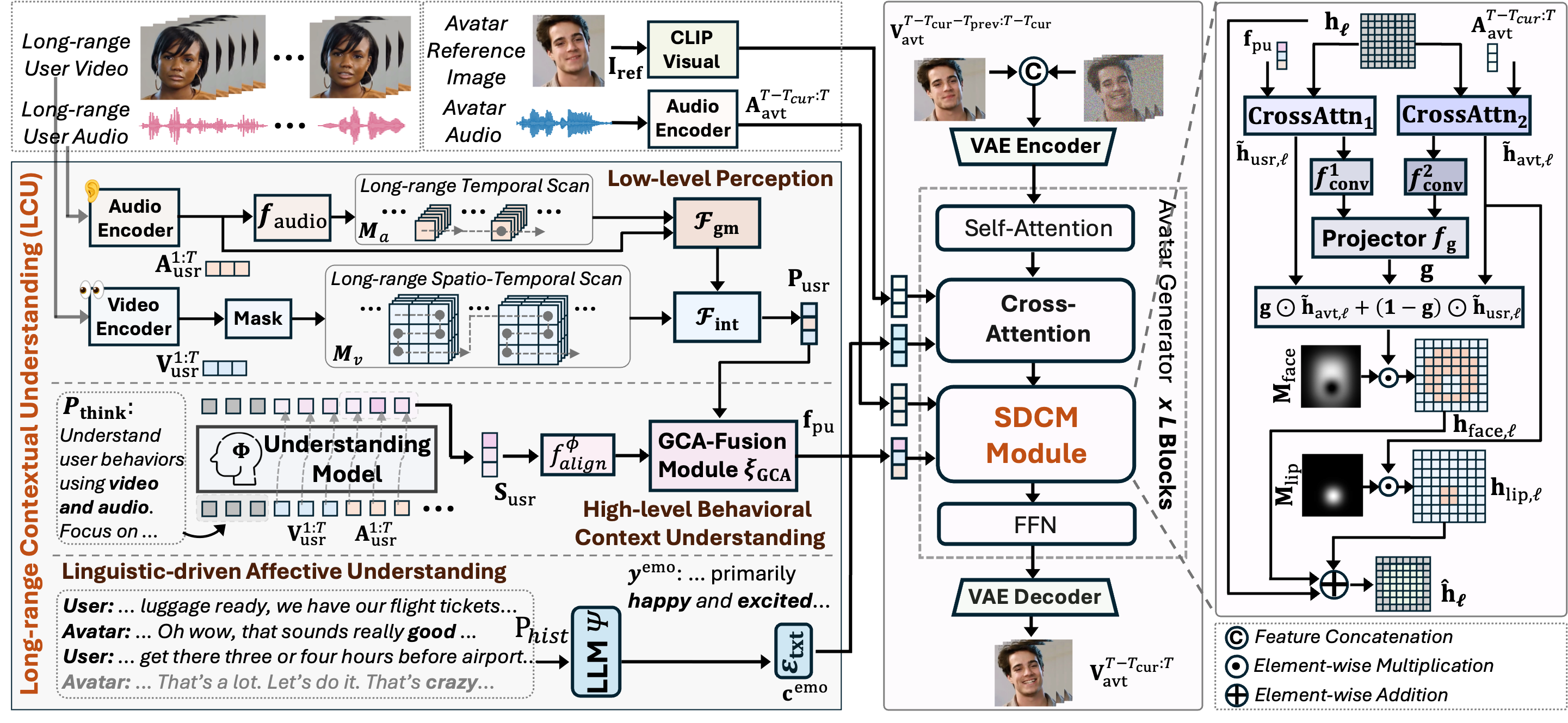}
  \caption{The pipeline of our proposed \textbf{ECHO}. The proposed \textbf{LCU} component starts by extracting user's long-range audio-visual behavioral features ($\mathbf{A}^{1:T}_{\text{usr}}$ and $\mathbf{V}^{1:T}_{\text{usr}}$) to perform low-level perception encoding along with high-level behavioral context understanding (resulting in perception-understanding representation $\textbf{f}_{\text{pu}}$). Then, linguistic dialogue context is leveraged to infer avatar's descriptive emotional state (resulting in embeddings $\mathbf{c}^{\text{emo}}$). Subsequently, two obtained representations are utilized as conditioning inputs for proposed block-wise \textbf{SDCM} module to guide avatar FBs generation. Extended pipeline details of avatar generator are provided in Appendix \textcolor{red}{2.1}.}
  \label{fig:pipline}
\end{figure}

As illustrated in Fig. \ref{fig:pipline}, our ECHO comprises two key components. The \textit{Long-range Contextual Understanding} (LCU) component (detailed in Sec. \ref{subsec:context_understanding}) begins by establishing understanding grounded in human user's long-range audio-visual behaviors (process denoted by $\mathcal{F}_{\text{BGU}}$). 
Specifically, low-level perception encoding is performed to capture fine-grained audio-visual dynamics ($\mathbf{P}_{\text{usr}}$); a multi-modal understanding model $\Phi$ is leveraged to distill hidden representations that captures high-level behavioral context cues.
Subsequently, these two key features are fused via our gated cross-attention-based (GCA)-fusion module $\xi_{\text{GCA}}$ to yield a unified perception-understanding representation $\mathbf{f}_{\text{pu}}$, formulated as:
\begin{equation}
    \mathbf{f}_{\text{pu}} = \mathcal{F}_{\text{BGU}}(\mathbf{V}^{1:T}_{\text{usr}}, \mathbf{A}^{1:T}_{\text{usr}}),
\end{equation}
where $\mathbf{f}_{\text{pu}}$ is utilized as conditioning guidance for generation of contextually appropriate avatar facial behaviors.
Complementarily, an LLM $\Psi$ is leveraged to facilitate linguistic-driven context understanding to further infer avatar’s descriptive emotional state, providing an explicit emotional guidance for enhancing the emotional rationality of generated avatar's FBs:
\begin{equation}
    y^{\text{emo}} = \Psi(\mathcal{P}_{\text{hist}}),
    \label{eq:emotion_prediction}
\end{equation}
Together, LCU aims to guide contextually appropriate and emotionally expressive facial behavior grounded in contextual understanding.

Furthermore, given the obtained $\mathbf{f}_{\text{pu}}$ and avatar’s self-audio $\mathbf{A}^{T-T_{\text{cur}}:T}_{\text{avt}}$, our \textit{Spatial-aware Decoupled Cross-Attention Modulation} (SDCM) component injects them into each generator block via region-wise decoupled cross-attentions: lip-region dynamics remain predominantly self-audio-driven to preserve articulation fidelity, while non-lip facial regions adaptively fuse user-conditioned and self-audio-driven influences via a learned gating to produce appropriately responsive yet self-audio-coordinated facial behaviors. SDCM is formulated as:
\begin{equation}
    \hat{\mathbf{h}}_\ell = \text{SDCM}_\ell(\mathbf{h}_\ell, \mathbf{f}_{\text{cu}}, \mathbf{A}^{T-T_{\text{cur}}:T}_{\text{avt}}),
\end{equation}
where $\mathbf{h}_\ell$ denotes the intermediate feature at block $\ell$ of the avatar generator (detailed in Sec. \ref{subsec:SDCM_module}). Finally, the proposed ECHO is trained with our two-stage flow-matching paradigm with HFA strategy (details in Sec. \ref{subsec:training_strategy}).

\subsection{Long-range Contextual Understanding (LCU)}
\label{subsec:context_understanding}

Our LCU starts by encoding the user’s long-range audio-visual behaviors into a perceptual representation, which captures fine-grained dynamics as low-level evidence for complementing subsequent high-level behavioral context understanding. 
Concretely, given user's audio features $\mathbf{A}^{1:T}_{\text{usr}}$ obtained with the per-frame context-window ($L_w$) scheme following \cite{xu2024hallo}, we first fuse window dimension via a learnable linear layer $f_{\text{audio}}:\mathbb{R}^{L_w\times D}\!\rightarrow\!\mathbb{R}^{D}$ (applied frame-wise over $T$, and $D$ denotes the feature dimensionality) to obtain frame-level embeddings, thereby enabling efficient temporal scanning with a $K$-layer Mamba\cite{gu2024mamba}-based module $\mathcal{M}_a$ for capturing long-range temporal dependency:
\begin{equation}
\begin{aligned}
    \tilde{\mathbf{A}}^{1:T}_{\text{usr},0}=f_{\text{audio}}(\mathbf{A}^{1:T}_{\text{usr}}),\quad
    \tilde{\mathbf{A}}^{1:T}_{\text{usr},k}=\tilde{\mathbf{A}}^{1:T}_{\text{usr},k-1}+
    \mathcal{M}_a^k\!\big(\mathrm{Norm}(\tilde{\mathbf{A}}^{1:T}_{\text{usr},k-1})\big),
\end{aligned}
\label{eq:audio_mamba}
\end{equation}
where $\text{Norm}(\cdot)$ denotes layer normalization. Here, each Mamba layer maintains a recurrent state updated by a scan along the temporal dimension, which acts as an implicit state-space memory for aggregating user's long-range audio cues.
To effectively incorporate such compressed long-range audio context into current-clip audio features spanning $T-T_{\text{cur}}:T$ while preserving fine-grained details, we leverage a gated residual modulation layer ($\mathcal{F}_{\text{gm}}$), computed as: 
\begin{equation}
\begin{aligned}
    \hat{\mathbf{A}}^{T-T_{\text{cur}}: T}_{\text{usr}} &= \mathbf{A}^{T - T_{\text{cur}}:T}_{\text{usr}} + \mathbf{g}_{\text{usr}} \odot \tanh(f_{\text{proj}}(\bar{\mathbf{A}}^{T-T_{\text{cur}}:T}_{\text{usr}})),
\end{aligned}
\label{eq:gated_injection}
\end{equation}
where $\bar{\mathbf{A}}^{T-T_{\text{cur}}:T}_{\text{usr}}$ is obtained by broadcasting $\tilde{\mathbf{A}}^{T-T_{\text{cur}}:T}_{\text{usr}}$ (embedding outputs from $\mathcal{M}_a$ spanning current $T_{\text{cur}}$-frame temporal clip) across context-window dimension to match the shape of $\mathbf{A}^{T - T_{\text{cur}}:T}_{\text{usr}}$, $f_{\text{proj}}$ denotes a linear layer, and $\mathbf{g}_{\text{usr}}$ denotes the element-wise gating tensor computed by a gated-fusion operator \cite{arevalo2017gated} conditioned on $\mathbf{A}^{T-T_{\text{cur}}:T}_{\text{usr}}$ and $\bar{\mathbf{A}}^{T-T_{\text{cur}}:T}_{\text{usr}}$.

Given visual behavior features $\mathbf{V}^{1:T}_{\text{usr}}$ derived from human user's head video, we first apply a facial mask $M_{\text{face}}$ to enforce visual representation learning specifically focusing on facial regions rather than irrelevant background, computed as:
\begin{equation}
\tilde{\mathbf{V}}^{1:T}_{\text{usr}} = \mathrm{Mask}(\mathbf{V}^{1:T}_{\text{usr}}, M_{\text{face}}),
\label{eq:face_mask}
\end{equation}
where $\mathrm{Mask}(\cdot)$ denotes feature selection under $M_{\text{face}}$ to filters out non-facial tokens.
Subsequently, a vision Mamba-based module $\mathcal{M}_v$ is leveraged for modeling long-range \textit{spatio-temporal} dependencies of visual cues, maintaining implicit state-space memory over both temporal and spatial dimensions: $\hat{\mathbf{V}}^{1:T}_{\text{usr}} = \mathcal{M}_v(\tilde{\mathbf{V}}^{1:T}_{\text{usr}})$.

Beyond long-range perception encoding that only captures fine-grained audio-visual dynamics, we leverage a multimodal understanding model $\Phi$ \cite{xu2025qwen2,xu2025qwen3} (i.e., which is large-scale pre-trained for multimodal understanding and reasoning over audio-video inputs) to distill high-level contextual representations $\mathbf{S}_{\text{usr}}$ grounded in user's long-range behaviors, empirically offering complementary semantic-level and affective-level cues for avatar FBs generation guidance. To steer $\Phi$ toward high-level thinking, we utilize an instruction prompt $\mathcal{P}_{\text{think}}$ (template provided in Appendix \textcolor{red}{3.3}) and extract last-layer hidden tokens:
\begin{equation}
    \mathbf{S}_{\text{usr}} = \Phi(\mathcal{P}_{\text{think}}, \mathbf{V}^{1:T}_{\text{usr}}, \mathbf{A}^{1:T}_{\text{usr}}),
    \label{eq:thinker_encoding}
\end{equation}

To make the perceptual encoding and high-level understanding complementary, we integrate the low-level perceptual features $\mathbf{P}_{\text{usr}}$ with the high-level representations $\mathbf{S}_{\text{usr}}$ via our gated cross-attention fusion module $\mathcal{\xi}_{\text{GCA}}$, so as to construct a unified perception-understanding representation, formulated as: 
\begin{equation}
\begin{aligned}
    \mathbf{f}_m &= \mathcal{A}_{\text{cross}}(\mathbf{P}_{\text{usr}}, f^{\Phi}_{\text{align}}(\mathbf{S}_{\text{usr}})) \cdot \tanh(\alpha) + \mathbf{P}_{\text{usr}}, \\
    \mathbf{f}_{\text{pu}} &= \mathcal{F}_\text{ffn}(\mathbf{f}_m) \cdot \tanh(\beta) + \mathbf{f}_m,
\end{aligned}
\label{eq:perception_understanding_fusion}
\end{equation}
where $\mathcal{A}_{\text{cross}}(\cdot, \cdot)$ denotes cross-attention; $f^{\Phi}_{\text{align}}$ represents the MLP-based projection module for representation dimensionality alignment; $\mathcal{F}_\text{ffn}$ denotes MLP-based feedforward module; $\alpha,\beta\in\mathbb{R}$ are learnable gating scalars. Finally, $\mathbf{f}_{\text{pu}}$ serves as conditioning feature to guide the avatar generator to produce contextually appropriate and emotionally rational facial behaviors.

Furthermore, as illustrated in Eq. \ref{eq:emotion_prediction} in Overview (Sec. \ref{sec:method}), the predicted emotional description $y^{\text{emo}}$ is further encoded via a text encoder $\mathcal{E}_{\text{txt}}$ \cite{raffel2020exploring}, formulated as: $\mathbf{c}^{\text{emo}} = \mathcal{E}_{\text{txt}}(y^{\text{emo}})$, and $\mathbf{c}^{\text{emo}}$ is subsequently injected as feature embeddings into the avatar generator via cross-attention (network details are provided in Appendix \textcolor{red}{2.1}) for guiding emotional appropriateness of avatar FBs.

\subsection{Spatial-aware Decoupled Cross-attention Modulation (SDCM)}
\label{subsec:SDCM_module}

We propose the SDCM module to jointly preserve lip articulation fidelity and enable contextually responsive non-lip facial behaviors. Specifically, given user behavior-grounded representation $\mathbf{f}_{\text{pu}}$ and the avatar's own audio $\mathbf{A}^{T-T_{\text{cur}}:T}_{\text{avt}}$, we utilize two decoupled cross-attention streams that each allow avatar generator's intermediate features $\mathbf{h}$ to independently attend to one conditioning branch:
\begin{equation}
\begin{aligned}
    \tilde{\mathbf{h}}_{\text{usr}} = \text{CrossAttn}_{1}(\mathbf{h}, \mathbf{f}_{\text{pu}}),\,\,\,\tilde{\mathbf{h}}_{\text{avt}} = \text{CrossAttn}_{2}(\mathbf{h}, \mathbf{A}^{T-T_{\text{cur}}:T}_{\text{avt}}),
\end{aligned}
\end{equation}
where $\tilde{\mathbf{h}}_{\text{usr}}$ captures user-driven contextual behavioral influence and $\tilde{\mathbf{h}}_{\text{avt}}$ encodes self-audio-driven dynamics.
Subsequently, following \cite{xu2024hallo}, we obtain pre-computed binary masks leveraging \cite{lugaresi2019mediapipe}: $\mathbf{M}_{\text{face}} \in \{0,1\}^{H' \times W'}$ indicating non-lip facial region and $\mathbf{M}_{\text{lip}} \in \{0,1\}^{H' \times W'}$ for lip region, where $H' \times W'$ denotes $\mathbf{h}$'s spatial resolution. Critically, lip region remains predominantly avatar self-audio-driven to preserve audio-visual lip-synchronization during speaking turns, formulated as: $\mathbf{h}_{\text{lip}} = \tilde{\mathbf{h}}_{\text{avt}} \odot \mathbf{M}_{\text{lip}}$. For non-lip facial regions, we compute a learned gating tensor $\mathbf{g}$ that adaptively interpolates between user-driven behavioral contextual influence and avatar self-audio-driven dynamics, producing contextually appropriate yet self-audio-coordinated facial behaviors, computed as:
\begin{equation}
\begin{aligned}
    \mathbf{h}_{\text{face}} &= \left( \mathbf{g} \odot \tilde{\mathbf{h}}_{\text{avt}} + (1 - \mathbf{g}) \odot \tilde{\mathbf{h}}_{\text{usr}} \right) \odot \mathbf{M}_{\text{face}}, \\
    \mathbf{g} &= \sigma\left(f_{\mathbf{g}}([f^1_{\text{conv}}(\tilde{\mathbf{h}}_{\text{usr}}); f^2_{\text{conv}}(\tilde{\mathbf{h}}_{\text{avt}})])\right),
    \label{eq:spatial_gate}
\end{aligned}
\end{equation}
where $[\cdot;\cdot]$ denotes channel-wise concatenation; $f^1_{\text{conv}}$ and $f^2_{\text{conv}}$ are learnable convolutional projectors; $f_{\mathbf{g}}$ denotes the gating projection; $\sigma$ denotes sigmoid activation; and $\odot$ denotes element-wise multiplication. The proposed SDCM module is applied at each block $\ell$ of $f_{\theta}$, with the block-wise output formulated as: $\hat{\mathbf{h}}_\ell = \mathbf{h}_\ell + \mathbf{h}_{\text{lip},\ell} + \mathbf{h}_{\text{face},\ell}$, enabling fine-grained hierarchical guidance.

\subsection{Two-Stage Training Paradigm}
\label{subsec:training_strategy}

Our training pipeline is structured into two sequential stages. Given the avatar-conditioned inputs (i.e., reference image $\mathbf{I}_{\text{ref}}$ and audio signals $\mathbf{A}^{T-T_{\text{cur}}:T}_{\text{avt}}$), the first stage aims to learn identity-consistent avatar head video generation with reliable self-audio-driven lip synchronization, by optimizing: $\mathcal{L}_{\text{Stage1}} = \mathcal{L}_{\text{flow}} + \lambda \mathcal{L}_{\text{pixel}}$,
where $\mathcal{L}_{\text{flow}}$ denotes standard flow-matching loss. $\lambda$ is a balancing coefficient, and $\mathcal{L}_{\text{pixel}}$ denotes an auxiliary pixel-space reconstruction loss (the details are provided in Appendix \textcolor{red}{2.2}) to further enhance perceptual fidelity of generated videos.

Stage~2 targets our core IHG objective. We freeze the Stage~1 trained avatar head generator as reference model ${f}_{\theta}^{\text{ref}}$ and adapt it by fine-tuning LoRA modules \cite{hu2022lora} on designated cross-attention layers (see Appendix \textcolor{red}{2.2} for layer settings), together with training the introduced LCU and SDCM components. The training objective combines flow-matching loss with a hierarchical feature alignment (HFA) regularizer: $\mathcal{L}_{\text{Stage2}} = \mathcal{L}_{\text{flow}} + \gamma \mathcal{L}_{\text{align}}$, where $\gamma$ is balancing coefficient, the HFA objective $\mathcal{L}_{\text{align}}$ enforces block-wise representational consistency with ${f}_{\theta}^{\text{ref}}$ through \textit{frame-level dynamic weighting}, formulated as:
\begin{equation}
    \mathcal{L}_{\text{align}} = \frac{1}{|\mathcal{S}|}
    \sum_{\ell \in \mathcal{S}} \sum_{i=T-T_{\text{cur}}}^{T}
    w(i)\cdot
    \overline{d(\mathbf{h}_\ell(i),
    \mathbf{h}_\ell^{\text{ref}}(i))},
\end{equation}
where $\overline{d(\cdot,\cdot)}$ denotes spatially averaged feature distance at frame $i$, $\mathcal{S}$ is a randomly sampled subset of blocks, and $\mathbf{h}_\ell(i)$ and $\mathbf{h}_\ell^{\text{ref}}(i)$ are intermediate features from Stage~2 adapted generator $f_{\theta}$ and Stage~1 reference model $f_{\theta}^{\text{ref}}$, respectively. The weight $w(i)$ is derived from frame-wise normalized short-time RMS energy of the avatar’s self-audio, serving as indicator of speech presence (details in Appendix \textcolor{red}{2.2}): speaking frames receive higher weights to preserve Stage~1 established talking fidelity (e.g., stable lip-synchronization), whereas listening frames are down-weighted to allow flexibility for learning contextually aware \textit{non-verbal} FBs generation that is absent in Stage~1 trained model.

\section{Experiments}
\label{sec:exp}

\subsection{Experimental settings}
\label{subsec:implement_details}

\noindent\textbf{Datasets:} We use the open-sourced \textit{Seamless Interaction} dataset \cite{agrawal2025seamless} for model training, which contains dyadic face-to-face conversational videos with synchronized audio for both participants. For video preprocessing, we detect and track faces using Face Alignment \cite{bulat2017far}, crop and resize each frame to $512 \times 512$ resolution, and resample all videos to 25 fps, finally yielding more than 300 hours of training data. For audio preprocessing, we extract audio features using Wav2Vec2 \cite{baevski2020wav2vec}. 

\noindent\textbf{Implementation details:} Training follows the two-stage paradigm described in Sec. \ref{subsec:training_strategy}.
Both stages employ LoRA \cite{hu2022lora} adaptation (rank $r{=}128$) on the avatar generator with a learning rate of $1 \times 10^{-4}$ and a base learning rate of $2 \times 10^{-5}$ for other trainable modules, using AdamW optimizer \cite{loshchilov2017decoupled} with a batch size of 16 on 16 NVIDIA A100 GPUs across 2 nodes. Further training and dataset details are provided in the Appendix \textcolor{red}{2.2} and \textcolor{red}{2.3}, respectively.

\noindent\textbf{Evaluation metrics:}
Our method is evaluated across four aspects: visual quality, lip-synchronization, facial motion quality, and contextual appropriateness of avatar emotional expressions. For visual quality, we adopt Fr\'{e}chet inception distance (\textbf{FID}) \cite{seitzer2020pytorch} and the Fr\'{e}chet video distance (\textbf{FVD}) \cite{unterthiner2018towards} for image-level and video-level generation quality respectively, while identity preservation is measured by cosine similarity of identity embeddings (\textbf{CSIM}) \cite{deng2019arcface} between generated frames and the avatar reference image. For lip synchronization, we employ \textbf{LSE-D} and \textbf{LSE-C} computed via SyncNet \cite{chung2016out} on generated avatar videos paired with corresponding audio, which measure lip-audio feature distance and synchronization confidence respectively.
For avatar facial motion quality, we extract FLAME blendshape parameters from generated videos using EMOCA \cite{danvevcek2022emoca}, decompose them into 50-d expression and 6-d pose components, and evaluate the generated motions from complementary perspectives following \cite{ng2022learning}, based on the rationale that \textit{high-quality interactive} facial motions should be not only accurate but also sufficiently diverse. Specifically, \textbf{MSE} measures point-wise reconstruction accuracy with respect to the ground-truth facial motion sequence, while \textbf{FD} evaluates motion realism at the distribution level by computing the Fr'{e}chet distance between the generated and ground-truth motion distributions, reflecting whether the generated facial motion sequences follow the distribution of natural facial dynamics. To assess motion diversity, \textbf{Var} quantifies the average temporal variance of generated facial motions across all feature dimensions within each sequence, capturing whether the avatar exhibits active and non-static dynamics, whereas \textbf{SID} \cite{ng2022learning} measures the diversity of generated facial motions by clustering the ground-truth facial motion sequences (including both expression and pose parameters) via $k$-means ($k=15,9$ for expression and pose, respectively) and computing the Shannon entropy of the predicted facial motion sequence's cluster assignment histogram; higher entropy indicates more diverse generated avatar FBs. Finally, \textbf{rPCC} evaluates user-avatar responsiveness by computing the absolute difference between the Pearson Correlation Coefficient (PCC) of ground-truth avatar facial motion with the user and that of the generated avatar facial motion with the user; a lower rPCC indicates that the generated avatar responds to the user in a manner more consistent with the ground-truth.
For \textit{Contextual appropriateness of avatar emotional expressions}: \textbf{E-Consis} denotes emotional consistency between Ground-truth and generated video pairs, scored on a 1--5 scale by Qwen3-Omni \cite{xu2025qwen3} upon analyzing and evaluating emotion category, intensity, and temporal alignment (the prompt template detailed in Appendix \textcolor{red}{3.5}); \textbf{E-Appro} represents human-rated (1--5 scale) emotional appropriateness of generated avatar head videos responsive to the corresponding real user videos.

\begin{figure}[t]
  \centering
  \includegraphics[width=0.9\linewidth]{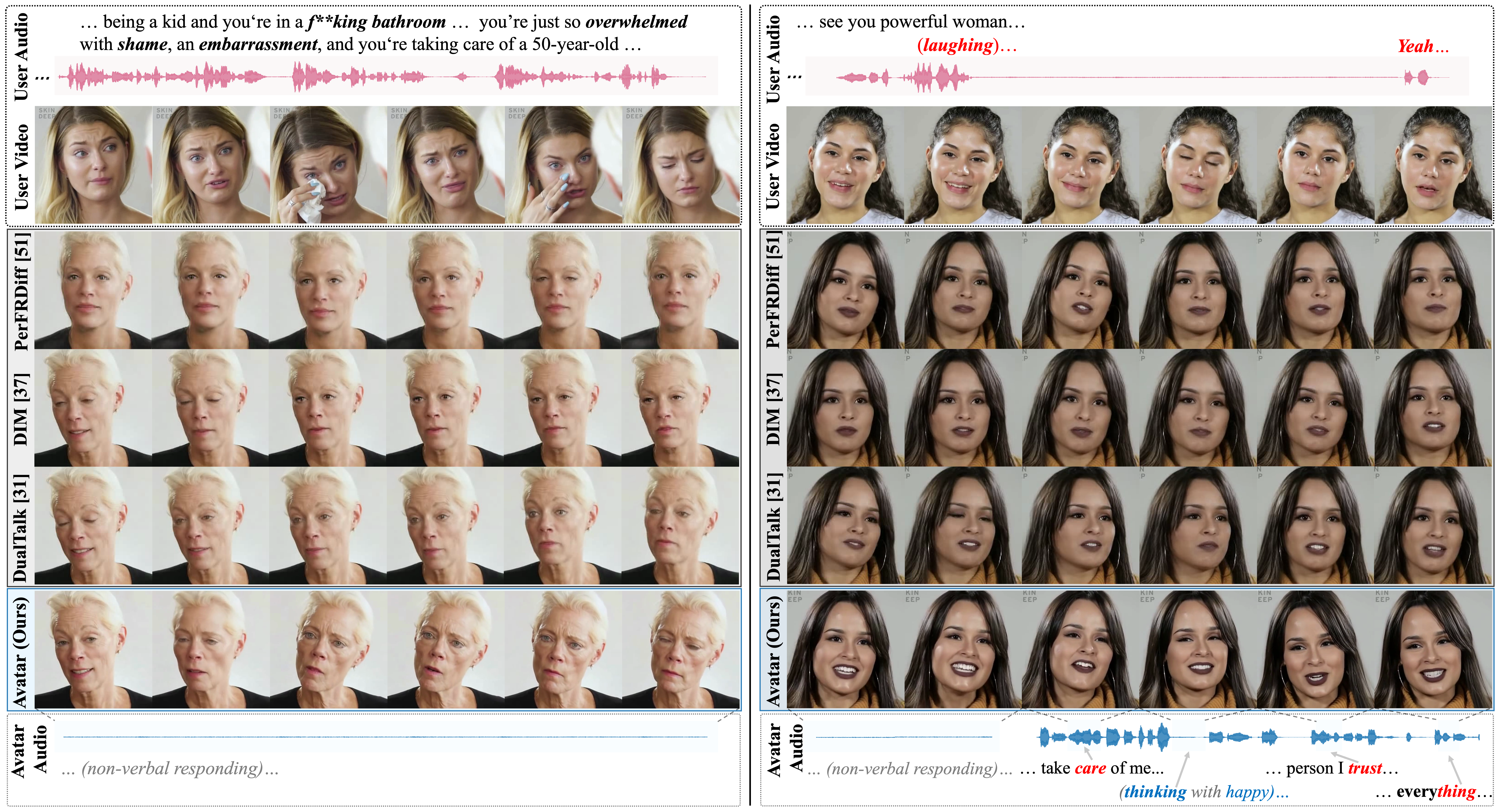}
  \caption{\textbf{Qualitative comparison} with open-sourced SOTA methods on IHG. Above two dyadic scenarios including active listening/speaking showcase that our proposed ECHO generates avatar head with more context-consistent and emotionally appropriate facial behaviors, and achieves lip articulation with better audio–visual alignment.}
  \label{fig:comparison}
\end{figure}

\subsection{Comparison Experiments}
We conduct quantitative evaluations across three tasks. We first compare with state-of-the-art (SOTA) IHG methods \cite{ki2026avatar,zhu2025infp,guo2025arig} on interactive head generation using both evaluation metrics described above and qualitative visualizations on the RealTalk \cite{geng2023affective} dataset. Following standard evaluation protocols \cite{ki2026avatar,zhu2025infp}, we additionally evaluate single-role generation capabilities without additional model adaptation or fine-tuning, including avatar listening head and talking face synthesis, to further validate our proposed ECHO's superiority.

\begin{table*}[t]
\centering
\caption{\textbf{Quantitative comparison} of \textit{interactive head generation} (IHG) on RealTalk \cite{geng2023affective} dataset. w/o LPE removes human user's long-range visual cues along with multimodal integration module $\mathcal{F}_{\text{int}}$ in LCU's low-level perception encoding. w/o HBCU retains LCU's perception encoding but discards $\Phi$-guided high-level behavior-grounded context understanding. w/o LAU disables dialogue-derived avatar descriptive affective state for guiding FBs generation.
\textbf{Bold}: best; \underline{Underline}: second-best.}
\label{tab:realtalk_results}
\resizebox{\textwidth}{!}{
\begin{tabular}{lcccccccc}
\toprule
& \textit{Reactiveness} & \multicolumn{2}{c}{\textit{Motion-Richness}} & \multicolumn{3}{c}{\textit{Visual-Quality}} & \multicolumn{2}{c}{\textit{Lip-Synchronization}} \\
Method & rPCC $\downarrow$ & SID $\uparrow$ & Var $\uparrow$ & FID $\downarrow$ & FVD $\downarrow$ & CSIM $\uparrow$ & LSE-D $\downarrow$ & LSE-C $\uparrow$ \\ \midrule
GT & 0.000 & 3.972 & 1.658 & - & - & 0.796 & 7.790 & 6.940 \\ \midrule
DIM~\cite{tran2024dim} & 0.186 & 1.083 & 1.206 & 26.290 & - & 0.843 & - & - \\
INFP~\cite{zhu2025infp} & - & 2.343 & 1.638 & 24.551 & \underline{159.000} & \underline{0.867} & \underline{8.027} & 6.536 \\
ARIG~\cite{guo2025arig} & \underline{0.125} & 2.428 & \textbf{2.397} & \underline{21.640} & - & \textbf{0.901} & - & - \\
AvatarForcing~\cite{ki2026avatar} & - & \underline{2.442} & \underline{1.734} & 24.328 & 170.874 & 0.833 & 8.060 & \textbf{6.723} \\
\textbf{Ours} & \textbf{0.109} & \textbf{2.631} & 1.543 & \textbf{19.877} & \textbf{152.349} & 0.825 & \textbf{7.899} & \underline{6.701} \\
\midrule
w/o LPE & 0.131 & 2.392 & 1.316 & 20.864 & 153.344 & 0.816 & 7.925 & 6.691 \\
w/o HBCU & 0.129 & 2.499 & 1.397 & 20.518 & 152.831  & 0.820 & 7.908 & 6.683 \\
w/o LAU & 0.124 & 2.558 & 1.486 & 20.083 & 152.909 & 0.827 & 7.914 & 6.689 \\
w/o SDCM & 0.119 & 2.601 & 1.519 & 21.857 & 154.910 & 0.809 & 8.009 & 6.572 \\
\bottomrule
\end{tabular}}
\end{table*}

\noindent\textbf{Comparison on Interactive Head Generation.} Table \ref{tab:realtalk_results} presents the quantitative comparison on the core IHG task, which comprehensively evaluates both speaking and listening capabilities under dyadic interactive settings. Our method achieves the best rPCC, indicating the strongest user-avatar motion synchronization and reflecting more natural reactive dynamics, and the highest SID, demonstrating that the generated FBs exhibit rich motion diversity rather than regressing toward uniform patterns (both improvements attributable to LCU component, whose long-range contextual understanding effectively promotes more reasonable and user-responsive facial behaviors).

For visual quality, our method attains the best FID and FVD, confirming superior image-level and video-level generation quality. For lip synchronization, our method achieves the best LSE-D and second-best LSE-C (within 0.02 of the best), validating that our designed SDCM architecture and two-stage training strategy effectively enhance lip-sync fidelity under dyadic interactive settings. We note that while our Var (1.543) does not rank highest, it is closer to the GT value (1.658) than the top-ranked result (2.397), indicating motion variability more consistent with the Ground-truth distribution, reflecting more plausible facial expressiveness.

Furthermore, \textbf{qualitative comparisons} highlight two consistent advantages of \textbf{ECHO}. In the first example of \cref{fig:comparison}, as the human user appears sad and even tearful with speaking explicit affective words (e.g., \textit{shame}, \textit{embarrassment}), ECHO generated a more concerned and even sad expression, whereas competitors remain largely in a faint smile with limited variation. This suggests that ECHO produces FBs that are more contextually appropriate and better aligned with human user's affective cues, further indicating the effectiveness of proposed LCU module. As shown in the second example, ECHO's generated avatar mirrors the user’s positive smile during listening and, after switching to speaking, shows clearer lip articulation with audio-visual alignment while even exhibiting brief gaze shifts suggestive of a “thinking” moment. The competitors show weaker emotional expressiveness and less reliable lip-synchronization. This indicates that ECHO better preserves lip-sync fidelity, which is empirically supported by our proposed SDCM module and two-stage training strategy.

\begin{figure}[htbp]
  \centering
  \includegraphics[width=0.95\linewidth]{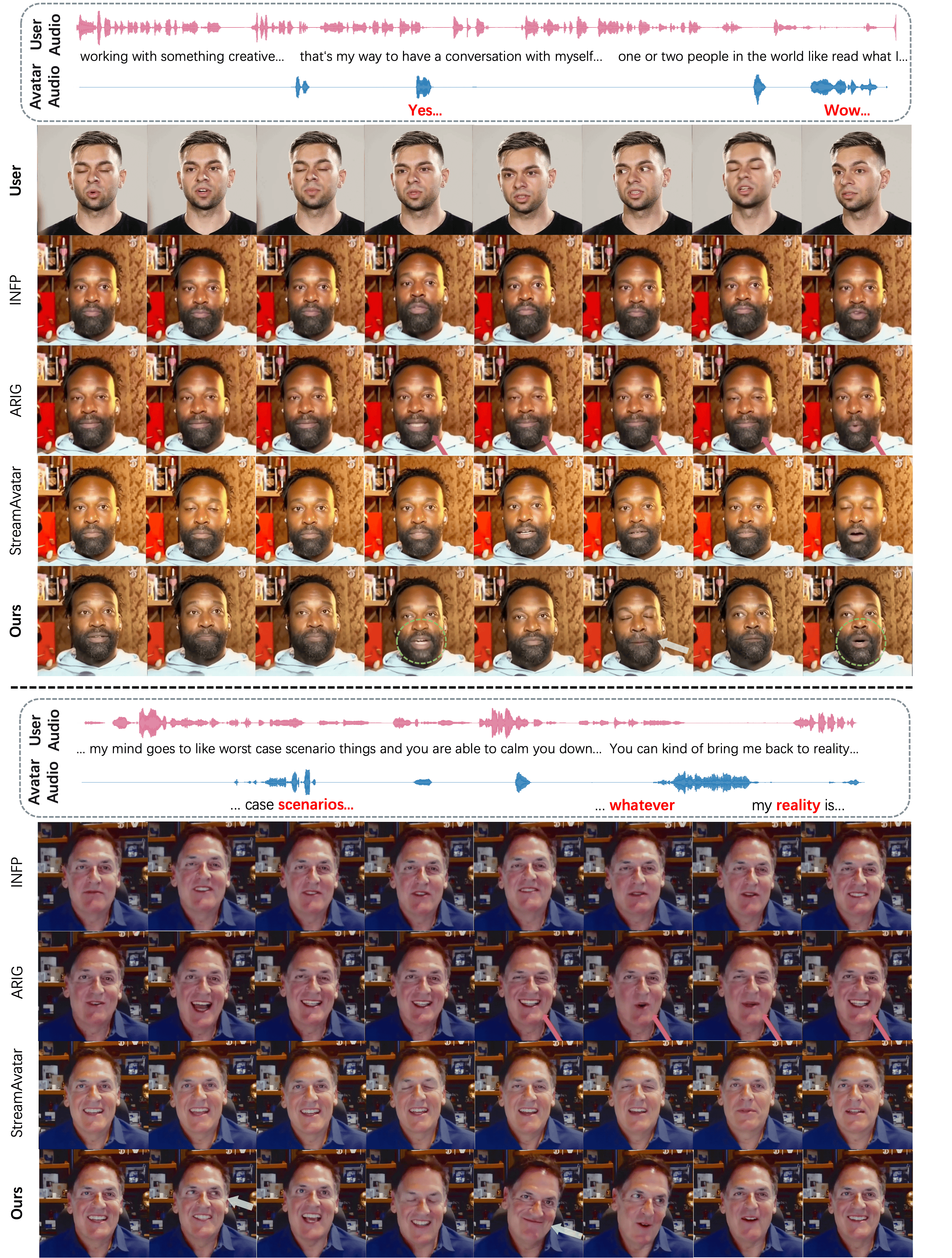}
  \caption{Qualitative comparison with state-of-the-art Interactive Head Generation (IHG) approaches, including INFP \cite{zhu2025infp}, ARIG \cite{guo2025arig}, and StreamAvatar \cite{sun2025streamavatar}, respectively. As all compared SOTA methods are not open-sourced, we follow their established visual comparison protocols and evaluate on two samples from the DyConv dataset introduced by INFP \cite{zhu2025infp}.}
  \label{fig:appendix_comparison1}
\end{figure}

\cref{fig:appendix_comparison1} presents an additional qualitative comparison against state-of-the-art IHG methods on samples from the DyConv dataset following the established evaluation protocol \cite{zhu2025infp}, demonstrating our ECHO's advantages in generating more nuanced and coherent avatar facial behaviors, evidenced by subtle lip compressions and natural gaze shifts during listening, as well as speech-synchronized head movements during speaking.\\

\noindent\textbf{Comparison on Single-role Avatar Head Generation.}
\textit{Talking Face Synthesis.} Following \cite{zhu2025infp,guo2025arig}, we randomly select fifty video samples from HDTF \cite{zhang2021flow} dataset as the test set. As shown in Table \ref{tab:hdtf_results}, our method achieved the best lip-sync scores in terms of both LSE-D and LSE-C, as well as the highest FID and FVD performance, outperforming both dedicated talking face synthesis methods \cite{xu2024hallo,zhang2023sadtalker,cui2025hallo3} and SOTA IHG approaches \cite{zhu2025infp,ki2026avatar,guo2025arig} with clear advantages, demonstrating the effectiveness of our ECHO in ensuring superior lip-synchronization and visual fidelity, which we attribute to our SDCM's architecture design and tailored training paradigm.

\noindent\textit{Listening Head Generation.} As shown in Table \ref{tab:vico_results}, in comparison to these SOTA IHG and listening head generation approaches, our method achieves the lowest FD on both generated expressions and poses, as well as the best rPCC on generated expressions, indicating that the generated facial motions exhibit the highest human-like realism among all compared methods. Our method also attains the best MSE on both expression and pose, and the highest SID on expression, demonstrating that the generated FBs closely align with Ground-truths while maintaining sufficient motion diversity. This together validate the superiority of our framework in avatar non-verbal FBs generation.

\begin{table}[h]
\centering
\small
\caption{Comparison with \textit{talking head generation} models on the HDTF \cite{zhang2021flow} dataset. $^\ast$ denotes the reproduced version reported in \cite{ki2026avatar}. \textbf{Bold}: best; \underline{Underline}: second-best.}
\label{tab:hdtf_results}
\begin{tabular}{lccccc}
\toprule
\multirow{2}{*}{Method} & \multicolumn{3}{c}{\textit{Visual Quality}} & \multicolumn{2}{c}{\textit{Lip Synchronization}} \\ \cmidrule(lr){2-4} \cmidrule(lr){5-6}
 & FID $\downarrow$ & FVD $\downarrow$ & CSIM $\uparrow$ & LSE-D $\downarrow$ & LSE-C $\uparrow$ \\ \midrule
Hallo~\cite{xu2024hallo} & 20.545 & 173.497 & 0.860 & 7.819 & 6.995 \\
SadTalker~\cite{zhang2023sadtalker} & 23.340 & 203.860 & 0.821 & 8.046 & 7.171 \\
Hallo3~\cite{cui2025hallo3} & 20.359 & 160.838 & {0.865} & 8.106 & 7.252 \\
INFP$^\ast$~\cite{zhu2025infp} & 27.155 & 187.977 & 0.840 & 7.810 & 7.325 \\
ARIG~\cite{guo2025arig} & \underline{18.320} & - & \textbf{0.876} & - & - \\
AvatarForcing~\cite{ki2026avatar} & 20.332 & \underline{149.798} & \underline{0.870} & \underline{7.700} & \underline{7.560} \\
\midrule
\textbf{Ours} & \textbf{16.888} & \textbf{144.859} & {0.854} & \textbf{7.553} & \textbf{7.704} \\ \bottomrule
\end{tabular}
\vspace{-0.1cm}
\end{table}

\begin{table}[h]
\centering
\small
\caption{Comparison with \textit{listening head generation} models on the ViCo \cite{zhou2022responsive} dataset. $^\dagger$ denotes metrics calculated via extracted 76-d 3DMM coefficients \cite{deng2019accurate}, whereas others utilize 56-d FLAME \cite{li2017learning} blendshape, thus presented for reference only. $^\ast$ refers to the reproduced version reported in \cite{ki2026avatar}. \textbf{Bold}: best; \underline{Underline}: second-best.}
\label{tab:vico_results}
\begin{tabular}{lcccccccccc}
\toprule
\multirow{2}{*}{Method} & \multicolumn{2}{c}{FD $\downarrow$} & \multicolumn{2}{c}{rPCC $\downarrow$} & \multicolumn{2}{c}{SID $\uparrow$} & \multicolumn{2}{c}{Var $\uparrow$} & \multicolumn{2}{c}{MSE $\downarrow$} \\ \cmidrule(lr){2-3} \cmidrule(lr){4-5} \cmidrule(lr){6-7} \cmidrule(lr){8-9} \cmidrule(lr){10-11}
 & Exp & Pose & Exp & Pose & Exp & Pose & Exp & Pose & Exp & Pose \\ \midrule
\rowcolor{gray!20} ARIG$^\dagger$~\cite{guo2025arig} & 18.39 & 0.06 & 0.05 & 0.01 & 4.82 & 3.94 & 2.91 & 0.17 & - & - \\
RLHG~\cite{zhou2022responsive} & 39.02 & 0.07 & 0.08 & \underline{0.02} & 3.62 & \underline{3.17} & 1.52 & 0.02 & 0.86 & \textbf{0.01} \\
L2L~\cite{ng2022learning} & 33.93 & 0.06 & 0.06 & 0.08 & 2.77 & 2.66 & 0.83 & 0.02 & 0.93 & \textbf{0.01} \\
DIM~\cite{tran2024dim} & 23.88 & 0.06 & 0.06 & 0.03 & \textbf{3.71} & 2.35 & 1.53 & 0.02 & 0.70 & \textbf{0.01} \\
DualTalk~\cite{peng2025dualtalk} & 22.27 & \underline{0.05} & - & - & - & - & - & - & 0.58 & \textbf{0.01} \\
INFP$^\ast$~\cite{zhu2025infp} & 17.52 & 0.07 & \textbf{0.01} & 0.07 & 2.19 & \textbf{3.20} & 2.10 & \underline{0.03} &
\underline{0.51} & \textbf{0.01} \\
AvatarForcing~\cite{ki2026avatar} & \underline{16.64} & \underline{0.05} & \textbf{0.01} & \textbf{0.01} & 3.12 & 3.00 & \textbf{2.80} & \underline{0.03} & - & - \\
\midrule
\textbf{Ours} & \textbf{15.67} & \textbf{0.04} & \textbf{0.01} & 0.07 & \textbf{3.71} & 2.93 & \underline{2.24} & 0.02 & \textbf{0.45} &
\textbf{0.01} \\ \bottomrule
\end{tabular}
\end{table}

\begin{table}[htbp]
  \centering
  \small
  \caption{Quantitative evaluation of emotional expressiveness in generated avatar head videos on RealTalk \cite{geng2023affective} dataset. Best results are \textbf{bold}, second best are \underline{underlined}.}
  \label{tab:emotion_results}
  \begin{tabular}{lcc}
    \toprule
    Method & \textbf{E-Consis} $\uparrow$ & \textbf{E-Appro} $\uparrow$ \\
    \midrule
    DIM & 3.271 & 3.390 \\
    DualTalk & \underline{3.422} & \underline{3.692} \\
    \textbf{Ours} & \textbf{3.798} & \textbf{4.073} \\
    \bottomrule
  \end{tabular}
  \vspace{-0.15cm}
\end{table}

\hfill \\
\noindent\textbf{Emotional Expressiveness Evaluation.}
Table \ref{tab:emotion_results} evaluates the emotional expressiveness of generated avatar head videos from two complementary perspectives: MLLM-based holistic assessment (E-Consis), and human subjective evaluation (E-Appro). For E-Consis, our method surpasses DualTalk and DIM, respectively, by a notable margin, demonstrating that our generated facial behaviors exhibit closer emotional alignment with Ground-truth videos. For human evaluation (E-Appro), we organized 25 participants to assess 50 generated avatar head videos (test videos randomly chosen from the RealTalk test set), our method receives the highest appropriateness rating, indicating that human evaluators perceive our model's generated FBs as more emotionally reasonable within the dyadic interactive context. In summary, consistent improvements indicate the effectiveness of proposed LCU component in enhancing emotional expressiveness and contextual appropriateness of generated avatar FBs. 

\textbf{Please refer to more comprehensive \href{https://xk0720.github.io/ECHO/}{visualization demos}} to fully demonstrate our ECHO's advantages across diverse conversational contexts, encompassing seamless interaction, listening, and speaking scenarios on the {RealTalk}, {HDTF} and {Seamless Interaction} test datasets.

\begin{figure}[t]
  \centering
  \includegraphics[width=0.85\linewidth]{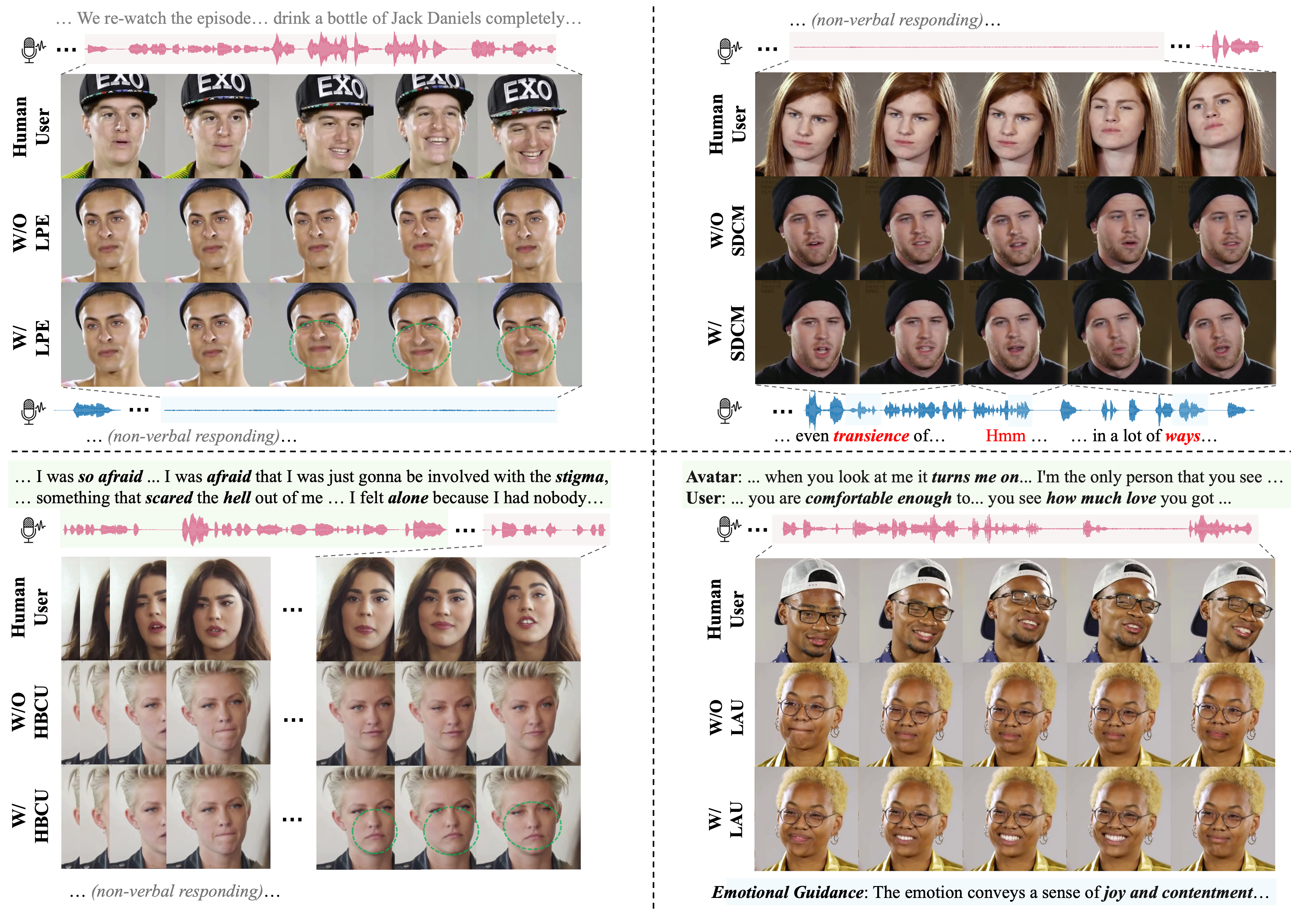}
  \caption{Ablation study: (a) Top-left: Ablation on low-level perception encoding by leveraging long-range visual cues; (b) Bottom Left: Ablation on high-level behavior-grounded context understanding; (c) Top-right: Ablation on SDCM module; (d) Bottom-right: Ablation on linguistic-driven affective understanding.}
  \label{fig:ablation}
\end{figure}

\subsection{Ablation Study}
To clearly attribute performance gains to individual designs, we report four component-level ablations and, for each, jointly analyze quantitative metrics in Table \ref{tab:realtalk_results} and qualitative results in Fig. \ref{fig:ablation}. Additional ablation study (e.g., on our two-stage training strategy) is provided in Appendix \textcolor{red}{3.1}.

\noindent\textbf{Effect of low-level perception encoding leveraging user visual cues (LPE).}
Removing human user's long-range visual cues (w/o LPE) together with multimodal integration $\mathcal{F}_{\text{int}}$ in LCU causes noticeable degradation in listening reactiveness (rPCC: 0.109$\rightarrow$0.131) and facial motion richness (SID: 2.631$\rightarrow$2.392 and Var: 1.543$\rightarrow$1.316). In the \textbf{top-left} example of Fig. \ref{fig:ablation}, the ablated model fails to mirror the user's smile and produces weakly expressive FBs, whereas the full model generates a timely matched smiling non-verbal response. These results indicate that, compared to audio-dominant conditioning, our LCU's perception encoding by leveraging user's long-range visual behaviors can effectively guide more user-synchronized and diverse facial behaviors.

\noindent\textbf{Effect of high-level behavior-grounded context understanding (HBCU).}
When retaining perception encoding but removing high-level understanding (w/o HBCU) in behavioral context, performance in avatar reactiveness and FBs diversity drops noticeably (reflected by degraded rPCC, SID and Var metrics). As in the \textbf{bottom-left} example (Fig. \ref{fig:ablation}), previous context conveys fear and distress through user's both utterances and facial behaviors, while the current clip appears neutral; the full model generates a serious expression consistent with the preceding context, whereas the ablated model defaults to a mild smile largely dominated by current temporal behavior clip. This contrast verifies the effectiveness of $\Phi$-guided behavior-grounded contextual understanding, which empirically contributes complementary semantic and affective cues from long-range user behaviors, in enabling the model to go beyond the current temporal clip for contextually appropriate FBs generation.

\noindent\textbf{Effect of the proposed SDCM module.}
\textit{w/o SDCM} replaces SDCM with a direct fusion strategy, where user-conditioned behavioral cues and avatar's self-audio features are fused into a single conditioning representation before injection into the avatar generator. As shown in the \textbf{top-right} example in Fig. \ref{fig:ablation}, incorporating SDCM reinforces avatar's self-audio dominance over lip-region during \textbf{\textit{speaking turns}} and yields more accurate lip-synchronization and audio-visual alignment, which is also consistent with improved LSE-D and LSE-C scores in Table \ref{tab:realtalk_results} as well as better FID and FVD metrics reflecting visual quality.

\noindent\textbf{Effect of linguistic-driven affective understanding (LAU).}
In the \textbf{bottom-right} example (Fig. \ref{fig:ablation}): informed by inferred emotional state description, the full model produces a visibly joyful and content facial behavior aligned with affective cues derived from the dialogue context and user's facial behavior, whereas disabling LAU yields a relatively neutral expression. Quantitatively, this behavioral neutralization is accompanied by reduced reactiveness (rPCC: 0.109$\rightarrow$0.124) and motion richness, verifying the plausibility of LAU-provided affective guidance and its effectiveness in enhancing avatar's emotional rationality.

\section{Conclusion}
We propose \textbf{ECHO}, a novel framework that enables contextually appropriate, emotionally rational facial behaviors and reliable lip synchronization in interactive head generation. ECHO introduces the LCU component that effectively provides long-range contextual understanding encompassing both behavior-grounded dynamics and linguistic-driven affective cues, and a block-wise SDCM design that prioritizes avatar self-audio for lip articulation while enriching non-lip FBs with user behavioral context, further complemented by our two-stage training paradigm with HFA strategy. Extensive experiments validate the effectiveness of proposed components and ECHO's superior IHG performance across emotional expressiveness, lip synchronization, and visual quality.

\noindent\textbf{Limitations and Future Work.} The main limitation in this work is that we have not yet explored optimizing flow-matching sampling efficiency, which would be beneficial for reducing inference latency and advancing toward real-time IHG. In future work, we plan to investigate two complementary directions to further enhance our framework: (1) advancing the block-wise causal attention architecture \cite{yin2025slow,chern2025livetalk} with KV caching to enable autoregressive streaming avatar generation, and (2) distilling multi-step sampling process into a few-step avatar generator \cite{yin2024improved,yin2024onestep,song2023consistency} to substantially reduce the required inference steps, jointly advancing our framework toward low-latency interactive deployment.

\clearpage
%
%
\bibliographystyle{splncs04}
\bibliography{main}

\begin{thebibliography}{10}
\providecommand{\url}[1]{\texttt{#1}}
\providecommand{\urlprefix}{URL }
\providecommand{\doi}[1]{https://doi.org/#1}

\bibitem{agrawal2025seamless}
Agrawal, V., Akinyemi, A., Alvero, K., Behrooz, M., Buffalini, J., Carlucci, F.M., Chen, J., Chen, J., Chen, Z., Cheng, S., et~al.: Seamless interaction: Dyadic audiovisual motion modeling and large-scale dataset. arXiv preprint arXiv:2506.22554  (2025)

\bibitem{arevalo2017gated}
Arevalo, J., Solorio, T., Montes-y G{\'o}mez, M., Gonz{\'a}lez, F.A.: Gated multimodal units for information fusion. arXiv preprint arXiv:1702.01992  (2017)

\bibitem{baevski2020wav2vec}
Baevski, A., Zhou, Y., Mohamed, A., Auli, M.: wav2vec 2.0: A framework for self-supervised learning of speech representations. Advances in neural information processing systems  \textbf{33},  12449--12460 (2020)

\bibitem{blanz2023morphable}
Blanz, V., Vetter, T.: A morphable model for the synthesis of 3d faces. In: Seminal Graphics Papers: Pushing the Boundaries, Volume 2, pp. 157--164 (2023)

\bibitem{boker2009effects}
Boker, S.M., Cohn, J.F., Theobald, B.J., Matthews, I., Brick, T.R., Spies, J.R.: Effects of damping head movement and facial expression in dyadic conversation using real--time facial expression tracking and synthesized avatars. Philosophical Transactions of the Royal Society B: Biological Sciences  \textbf{364}(1535),  3485--3495 (2009)

\bibitem{bulat2017far}
Bulat, A., Tzimiropoulos, G.: How far are we from solving the 2d \& 3d face alignment problem?(and a dataset of 230,000 3d facial landmarks). In: Proceedings of the IEEE international conference on computer vision. pp. 1021--1030 (2017)

\bibitem{cai2025towards}
Cai, Y., Chu, X., Gao, X., Gong, S., Huang, Y., Kang, C., Li, K., Liu, H., Liu, R., Liu, Y., et~al.: Towards interactive intelligence for digital humans. arXiv preprint arXiv:2512.13674  (2025)

\bibitem{chartrand1999chameleon}
Chartrand, T.L., Bargh, J.A.: The chameleon effect: The perception--behavior link and social interaction. Journal of personality and social psychology  \textbf{76}(6), ~893 (1999)

\bibitem{chern2025livetalk}
Chern, E., Hu, Z., Tang, B., Su, J., Chern, S., Deng, Z., Liu, P.: Livetalk: Real-time multimodal interactive video diffusion via improved on-policy distillation. arXiv preprint arXiv:2512.23576  (2025)

\bibitem{chung2016out}
Chung, J.S., Zisserman, A.: Out of time: automated lip sync in the wild. In: Asian conference on computer vision. pp. 251--263. Springer (2016)

\bibitem{cui2025hallo3}
Cui, J., Li, H., Zhan, Y., Shang, H., Cheng, K., Ma, Y., Mu, S., Zhou, H., Wang, J., Zhu, S.: Hallo3: Highly dynamic and realistic portrait image animation with video diffusion transformer. In: Proceedings of the Computer Vision and Pattern Recognition Conference. pp. 21086--21095 (2025)

\bibitem{danvevcek2022emoca}
Dan{\v{e}}{\v{c}}ek, R., Black, M.J., Bolkart, T.: Emoca: Emotion driven monocular face capture and animation. In: Proceedings of the IEEE/CVF conference on computer vision and pattern recognition. pp. 20311--20322 (2022)

\bibitem{deng2019arcface}
Deng, J., Guo, J., Xue, N., Zafeiriou, S.: Arcface: Additive angular margin loss for deep face recognition. In: Proceedings of the IEEE/CVF conference on computer vision and pattern recognition. pp. 4690--4699 (2019)

\bibitem{deng2019accurate}
Deng, Y., Yang, J., Xu, S., Chen, D., Jia, Y., Tong, X.: Accurate 3d face reconstruction with weakly-supervised learning: From single image to image set. In: Proceedings of the IEEE/CVF conference on computer vision and pattern recognition workshops. pp.~0--0 (2019)

\bibitem{geng2023affective}
Geng, S., Teotia, R., Tendulkar, P., Menon, S., Vondrick, C.: Affective faces for goal-driven dyadic communication. arXiv preprint arXiv:2301.10939  (2023)

\bibitem{gu2024mamba}
Gu, A., Dao, T.: Mamba: Linear-time sequence modeling with selective state spaces. In: First conference on language modeling (2024)

\bibitem{guo2025deepseek}
Guo, D., Yang, D., Zhang, H., Song, J., Zhang, R., Xu, R., Zhu, Q., Ma, S., Wang, P., Bi, X., et~al.: Deepseek-r1: Incentivizing reasoning capability in llms via reinforcement learning. arXiv preprint arXiv:2501.12948  (2025)

\bibitem{guo2024liveportrait}
Guo, J., Zhang, D., Liu, X., Zhong, Z., Zhang, Y., Wan, P., Zhang, D.: Liveportrait: Efficient portrait animation with stitching and retargeting control. arXiv preprint arXiv:2407.03168  (2024)

\bibitem{guo2025arig}
Guo, Y., Liu, X., Zhen, C., Yan, P., Wei, X.: Arig: Autoregressive interactive head generation for real-time conversations. In: Proceedings of the IEEE/CVF International Conference on Computer Vision (ICCV). pp. 12956--12965 (October 2025)

\bibitem{hatfield1993emotional}
Hatfield, E., Cacioppo, J.T., Rapson, R.L.: Emotional contagion. Current directions in psychological science  \textbf{2}(3),  96--100 (1993)

\bibitem{hu2022lora}
Hu, E.J., Shen, Y., Wallis, P., Allen-Zhu, Z., Li, Y., Wang, S., Wang, L., Chen, W.: Lo{RA}: Low-rank adaptation of large language models. In: International Conference on Learning Representations (2022), \url{https://openreview.net/forum?id=nZeVKeeFYf9}

\bibitem{jiang2025omnihuman}
Jiang, J., Zeng, W., Zheng, Z., Yang, J., Liang, C., Liao, W., Liang, H., Zhang, Y., Gao, M.: Omnihuman-1.5: Instilling an active mind in avatars via cognitive simulation. arXiv preprint arXiv:2508.19209  (2025)

\bibitem{ki2026avatar}
Ki, T., Jang, S., Jo, J., Yoon, J., Hwang, S.J.: Avatar forcing: Real-time interactive head avatar generation for natural conversation. arXiv preprint arXiv:2601.00664  (2026)

\bibitem{ki2025float}
Ki, T., Min, D., Chae, G.: Float: Generative motion latent flow matching for audio-driven talking portrait. In: Proceedings of the IEEE/CVF International Conference on Computer Vision. pp. 14699--14710 (2025)

\bibitem{li2017learning}
Li, T., Bolkart, T., Black, M.J., Li, H., Romero, J.: Learning a model of facial shape and expression from 4d scans. ACM Trans. Graph.  \textbf{36}(6),  194--1 (2017)

\bibitem{lin2025omnihuman}
Lin, G., Jiang, J., Yang, J., Zheng, Z., Liang, C., Zhang, Y., Liu, J.: Omnihuman-1: Rethinking the scaling-up of one-stage conditioned human animation models. In: Proceedings of the IEEE/CVF International Conference on Computer Vision. pp. 13847--13858 (2025)

\bibitem{liu2024customlistener}
Liu, X., Guo, Y., Zhen, C., Li, T., Ao, Y., Yan, P.: Customlistener: Text-guided responsive interaction for user-friendly listening head generation. In: Proceedings of the IEEE/CVF Conference on Computer Vision and Pattern Recognition. pp. 2415--2424 (2024)

\bibitem{loshchilov2017decoupled}
Loshchilov, I., Hutter, F.: Decoupled weight decay regularization. arXiv preprint arXiv:1711.05101  (2017)

\bibitem{lugaresi2019mediapipe}
Lugaresi, C., Tang, J., Nash, H., McClanahan, C., Uboweja, E., Hays, M., Zhang, F., Chang, C.L., Yong, M.G., Lee, J., et~al.: Mediapipe: A framework for building perception pipelines. arXiv preprint arXiv:1906.08172  (2019)

\bibitem{luo2024reactface}
Luo, C., Song, S., Xie, W., Spitale, M., Ge, Z., Shen, L., Gunes, H.: Reactface: Online multiple appropriate facial reaction generation in dyadic interactions. IEEE Transactions on Visualization and Computer Graphics  \textbf{31}(9),  6190--6207 (2024)

\bibitem{ng2022learning}
Ng, E., Joo, H., Hu, L., Li, H., Darrell, T., Kanazawa, A., Ginosar, S.: Learning to listen: Modeling non-deterministic dyadic facial motion. In: Proceedings of the IEEE/CVF Conference on Computer Vision and Pattern Recognition. pp. 20395--20405 (2022)

\bibitem{ng2023can}
Ng, E., Subramanian, S., Klein, D., Kanazawa, A., Darrell, T., Ginosar, S.: Can language models learn to listen? In: Proceedings of the IEEE/CVF International Conference on Computer Vision. pp. 10083--10093 (2023)

\bibitem{peebles2023scalable}
Peebles, W., Xie, S.: Scalable diffusion models with transformers. In: Proceedings of the IEEE/CVF international conference on computer vision. pp. 4195--4205 (2023)

\bibitem{peng2025dualtalk}
Peng, Z., Fan, Y., Wu, H., Wang, X., Liu, H., He, J., Fan, Z.: Dualtalk: Dual-speaker interaction for 3d talking head conversations. In: Proceedings of the Computer Vision and Pattern Recognition Conference. pp. 21055--21064 (2025)

\bibitem{raffel2020exploring}
Raffel, C., Shazeer, N., Roberts, A., Lee, K., Narang, S., Matena, M., Zhou, Y., Li, W., Liu, P.J.: Exploring the limits of transfer learning with a unified text-to-text transformer. Journal of machine learning research  \textbf{21}(140),  1--67 (2020)

\bibitem{rombach2021highresolution}
Rombach, R., Blattmann, A., Lorenz, D., Esser, P., Ommer, B.: High-resolution image synthesis with latent diffusion models (2021)

\bibitem{seitzer2020pytorch}
Seitzer, M.: pytorch-fid: Fid score for pytorch (2020)

\bibitem{song2025react}
Song, S., Spitale, M., Kong, X., Zhu, H., Luo, C., Palmero, C., Barquero, G., Escalera, S., Valstar, M., Daoudi, M., et~al.: React 2025: the third multiple appropriate facial reaction generation challenge. In: Proceedings of the 33rd ACM International Conference on Multimedia. pp. 13979--13984 (2025)

\bibitem{song2023consistency}
Song, Y., Dhariwal, P., Chen, M., Sutskever, I.: Consistency models  (2023)

\bibitem{sun2025streamavatar}
Sun, Z., Peng, Z., Ma, Y., Chen, Y., Zhou, Z., Zhou, Z., Zhang, G., Zhang, Y., Zhou, Y., Lu, Q., et~al.: Streamavatar: Streaming diffusion models for real-time interactive human avatars. arXiv preprint arXiv:2512.22065  (2025)

\bibitem{tran2024dim}
Tran, M., Chang, D., Siniukov, M., Soleymani, M.: Dim: Dyadic interaction modeling for social behavior generation. In: European Conference on Computer Vision. pp. 484--503. Springer (2024)

\bibitem{unterthiner2018towards}
Unterthiner, T., Van~Steenkiste, S., Kurach, K., Marinier, R., Michalski, M., Gelly, S.: Towards accurate generative models of video: A new metric \& challenges. arXiv preprint arXiv:1812.01717  (2018)

\bibitem{xu2025qwen2}
Xu, J., Guo, Z., He, J., Hu, H., He, T., Bai, S., Chen, K., Wang, J., Fan, Y., Dang, K., et~al.: Qwen2. 5-omni technical report. arXiv preprint arXiv:2503.20215  (2025)

\bibitem{xu2025qwen3}
Xu, J., Guo, Z., Hu, H., Chu, Y., Wang, X., He, J., Wang, Y., Shi, X., He, T., Zhu, X., et~al.: Qwen3-omni technical report. arXiv preprint arXiv:2509.17765  (2025)

\bibitem{xu2024hallo}
Xu, M., Li, H., Su, Q., Shang, H., Zhang, L., Liu, C., Wang, J., Yao, Y., Zhu, S.: Hallo: Hierarchical audio-driven visual synthesis for portrait image animation. arXiv preprint arXiv:2406.08801  (2024)

\bibitem{yin2024improved}
Yin, T., Gharbi, M., Park, T., Zhang, R., Shechtman, E., Durand, F., Freeman, B.: Improved distribution matching distillation for fast image synthesis. Advances in neural information processing systems  \textbf{37},  47455--47487 (2024)

\bibitem{yin2024onestep}
Yin, T., Gharbi, M., Zhang, R., Shechtman, E., Durand, F., Freeman, W.T., Park, T.: One-step diffusion with distribution matching distillation. In: CVPR (2024)

\bibitem{yin2025slow}
Yin, T., Zhang, Q., Zhang, R., Freeman, W.T., Durand, F., Shechtman, E., Huang, X.: From slow bidirectional to fast autoregressive video diffusion models. In: Proceedings of the IEEE/CVF Conference on Computer Vision and Pattern Recognition. pp. 22963--22974 (2025)

\bibitem{zhang2023sadtalker}
Zhang, W., Cun, X., Wang, X., Zhang, Y., Shen, X., Guo, Y., Shan, Y., Wang, F.: Sadtalker: Learning realistic 3d motion coefficients for stylized audio-driven single image talking face animation. In: Proceedings of the IEEE/CVF conference on computer vision and pattern recognition. pp. 8652--8661 (2023)

\bibitem{zhang2021flow}
Zhang, Z., Li, L., Ding, Y., Fan, C.: Flow-guided one-shot talking face generation with a high-resolution audio-visual dataset. In: Proceedings of the IEEE/CVF conference on computer vision and pattern recognition. pp. 3661--3670 (2021)

\bibitem{zhou2025interactive}
Zhou, M., Bai, Y., Zhang, W., Yao, T., Zhao, T.: Interactive conversational head generation. IEEE Transactions on Pattern Analysis and Machine Intelligence  (2025)

\bibitem{zhou2022responsive}
Zhou, M., Bai, Y., Zhang, W., Yao, T., Zhao, T., Mei, T.: Responsive listening head generation: a benchmark dataset and baseline. In: European conference on computer vision. pp. 124--142. Springer (2022)

\bibitem{zhou2020makelttalk}
Zhou, Y., Han, X., Shechtman, E., Echevarria, J., Kalogerakis, E., Li, D.: Makelttalk: speaker-aware talking-head animation. ACM Transactions On Graphics (TOG)  \textbf{39}(6),  1--15 (2020)

\bibitem{zhu2025perreactor}
Zhu, H., Kong, X., Xie, W., Huang, X., He, X., Liu, L., Shen, L., Zhang, W., Gunes, H., Song, S.: Perreactor: Offline personalised multiple appropriate facial reaction generation. In: Proceedings of the AAAI Conference on Artificial Intelligence. vol.~39, pp. 1665--1673 (2025)

\bibitem{zhu2024perfrdiff}
Zhu, H., Kong, X., Xie, W., Huang, X., Shen, L., Liu, L., Gunes, H., Song, S.: Perfrdiff: Personalised weight editing for multiple appropriate facial reaction generation. In: Proceedings of the 32nd ACM International Conference on Multimedia. pp. 9495--9504 (2024)

\bibitem{zhu2025infp}
Zhu, Y., Zhang, L., Rong, Z., Hu, T., Liang, S., Ge, Z.: Infp: Audio-driven interactive head generation in dyadic conversations. In: Proceedings of the Computer Vision and Pattern Recognition Conference. pp. 10667--10677 (2025)

\end{thebibliography}

\end{document}